\documentclass{article}
\usepackage{spconf,amsmath,graphicx}

\title{UNCERTAINTY-BASED METHOD FOR IMPROVING POORLY LABELED SEGMENTATION DATASETS}
\name{Ekaterina Redekop, Alexey Chernyavskiy}
\address{Philips AI Research, Moscow, Russia}
\begin{document}
%
\maketitle
\begin{abstract}
The success of modern deep learning algorithms for image segmentation heavily depends on the availability of large datasets with clean pixel-level annotations (masks), where the objects of interest are accurately delineated. Lack of time and expertise during data annotation leads to incorrect boundaries and label noise. It is known that deep convolutional neural networks (DCNNs) can memorize even completely random labels, resulting in poor accuracy. We propose a framework to train binary segmentation DCNNs using sets of unreliable pixel-level annotations. Erroneously labeled pixels are identified based on the estimated aleatoric uncertainty of the segmentation and are relabeled to the true value. 
\end{abstract}
\section{Introduction}
\label{sec:intro}

In many clinical applications medical image segmentation is an essential step that foregoes quantitative analysis of clinical features and diagnosis. DCNNs showed state-of-the-art performance for segmentation tasks in case when the training sets are large, representative and correctly annotated. In case of noisy annotations, DCNNs easily overfits to them during training, resulting in mediocre performance on the test sets \cite{zhang2016understanding}.  Compared to consumer photography, obtaining large datasets consisting of accurately labeled medical images is expensive and much more time-consuming. It is therefore essential to devise methods that will produce stable results while training on relatively cheap and unreliable annotations. 

In image segmentation, noisy labels refer to inaccuracies or ambiguities of mask boundaries. The major sources of label noise include inter-observer variability due to human subjectivity, random mistakes made by human annotators, and errors in computer-generated labels.
  

There exist several solutions that deal with, or account for label noise in deep learning \cite{tajbakhsh2020embracing}. Most of these strategies were developed for the classification task, where a single label corresponds to each data sample, e.g. a whole image. When it is meaningful and feasible, these strategies are transformed to solve image segmentation problem with pixel-level labels. For example, the idea of image reweighting proposed in \cite{ren2018learning} is used by \cite{mirikharaji2019learning} to solve the image segmentation task for dermatoscopy using synthesized noisy masks of skin lesions.

The presence of noisy ground truth data during training leads to uncertainties in model predictions at test time. Several tools for uncertainty estimation have recently been proposed. Highly uncertain samples or regions can be referred to an expert for examination; label correction can also be done automatically  \cite{kohler2019uncertainty}. The authors used uncertainty estimation for iterative detection and filtering of noisy labels, which showed promising performance for the image classification task. In this work we extend this approach to the task of binary image segmentation by using pixel-level uncertainty estimation to detect and relabel noisy ground truth masks for dermatoscopy and liver CT data.  

We believe that the novelty and scientific contribution of our study is  three-fold. First, we improve binary segmentation quality on a test set without overfitting to data inaccuracies present in the ground truth masks during training. Second, our relabeling algorithm does not depend on the availability or knowledge of subsets of trained data that are guaranteed to be accurate. Finally, the proposed solution allows to avoid discarding noisy ground truth data. On the contrary, the effect of this work is to post-process noisy datasets and make them them more accurate. In this way, it becomes possible for the annotators to spend less time and effort during annotation of images, since an automatic correction of masks can be performed.
\begin{figure*}[!htb]
\centering
\includegraphics[width=0.8\textwidth]
{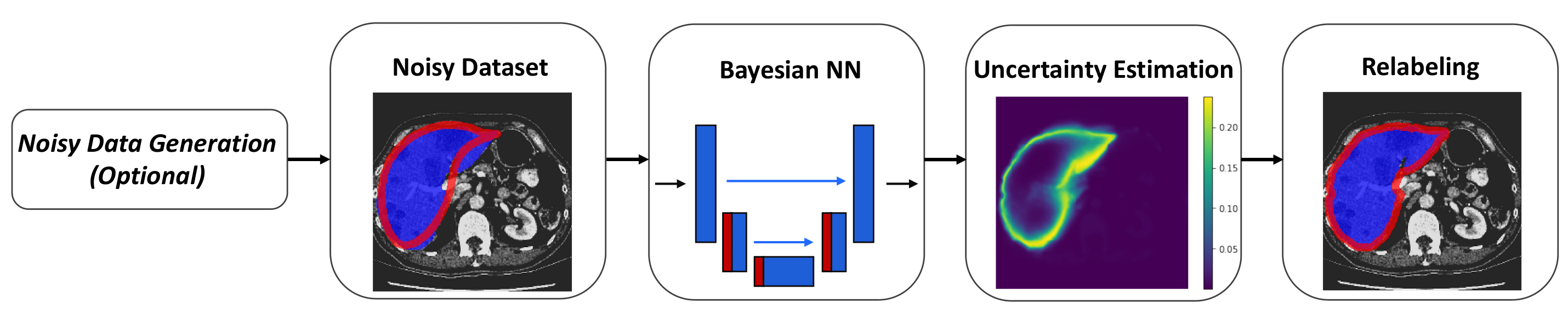}
\caption{Detection and relabeling of noisy ground truth labels for binary image segmentation\label{fig:pipeline}} 
\end{figure*}

\begin{figure*}[!htb]
\centering
\includegraphics[width=0.7\textwidth]
{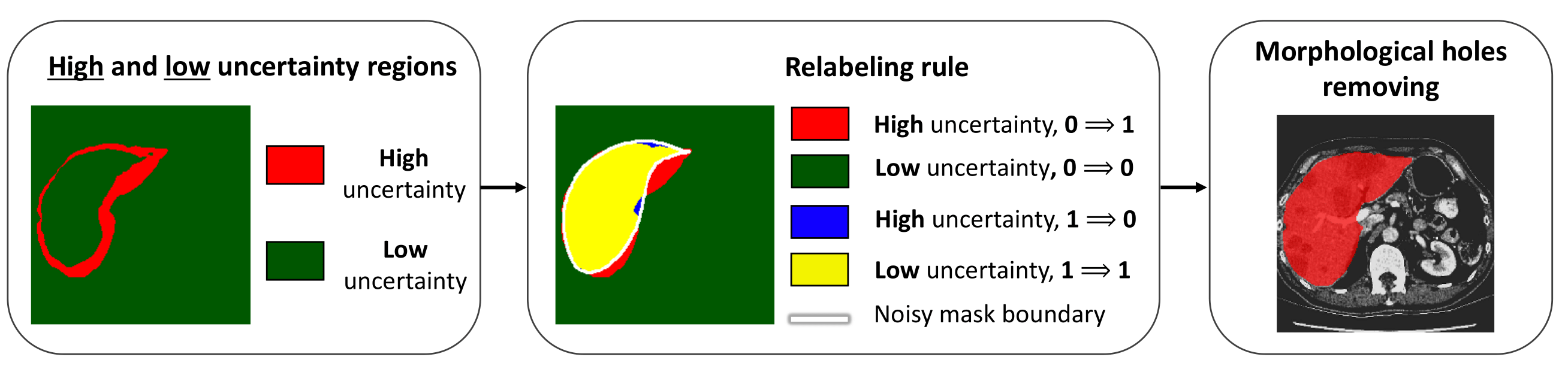}
\caption{Details of the relabeling step \label{fig:pipeline2}} 
\end{figure*}

\section{Methodology}
\label{sec:meto}
In this section we describe the proposed method of uncertainty-based noisy labels detection, and a way of relabeling  them, for the task of binary (object vs. background) image segmentation. The framework is illustrated in Figure~\ref{fig:pipeline} and described next.
\subsection{Noisy data generation}
\label{ssec:subhead}
\begin{figure}[htb]

\begin{minipage}[b]{.3\linewidth}
  \centering
  \centerline{\includegraphics[width=2.8cm]{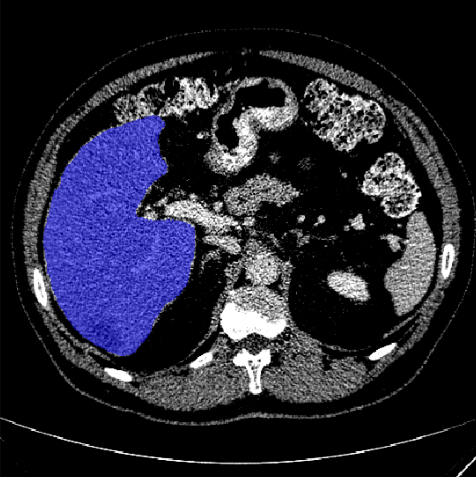}}
  \centerline{(a)}\medskip
\end{minipage}
\hfill
\begin{minipage}[b]{0.3\linewidth}
  \centering
  \centerline{\includegraphics[width=2.8cm]{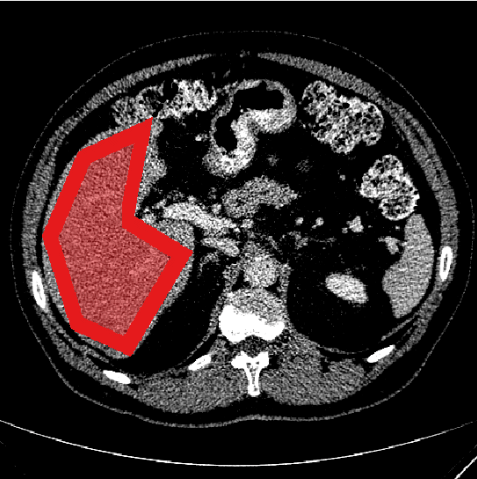}}
  \centerline{(b)}\medskip
\end{minipage}
\hfill
\begin{minipage}[b]{0.3\linewidth}
  \centering
  \centerline{\includegraphics[width=2.8cm]{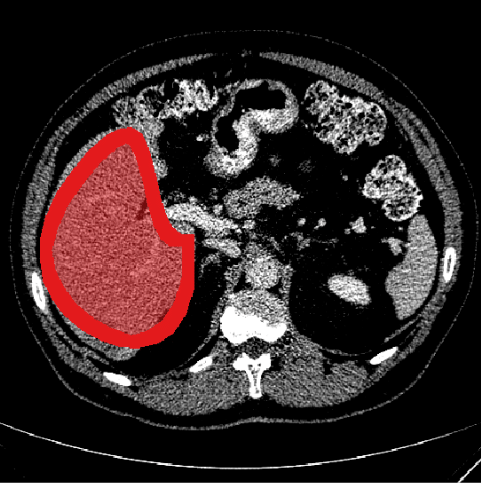}}
  \centerline{(c)}\medskip
\end{minipage}
\caption{(a) Ground truth mask, (b) low-vertex polygon approximation, (c) smooth curve approximation}
\label{fig:noisy_data}
\end{figure}

It is assumed that the segmentation masks given in the training set are inherently noisy, otherwise clean ground truth masks should be artificially deteriorated. One of the approaches to generate noisy ground truth annotations is to consider an object boundary as a closed polygon, then reduce the number of its vertices, as was proposed in \cite{mirikharaji2019learning}, and then either connect the remaining vertices by line segments, or fit a smooth curve to them (Figure~\ref{fig:noisy_data}).

\subsection{Uncertainty estimation for DCNNs}
\label{ssec:sub_unc}

\begin{figure*}[!htb]
\centering
\includegraphics[width=1\textwidth]
{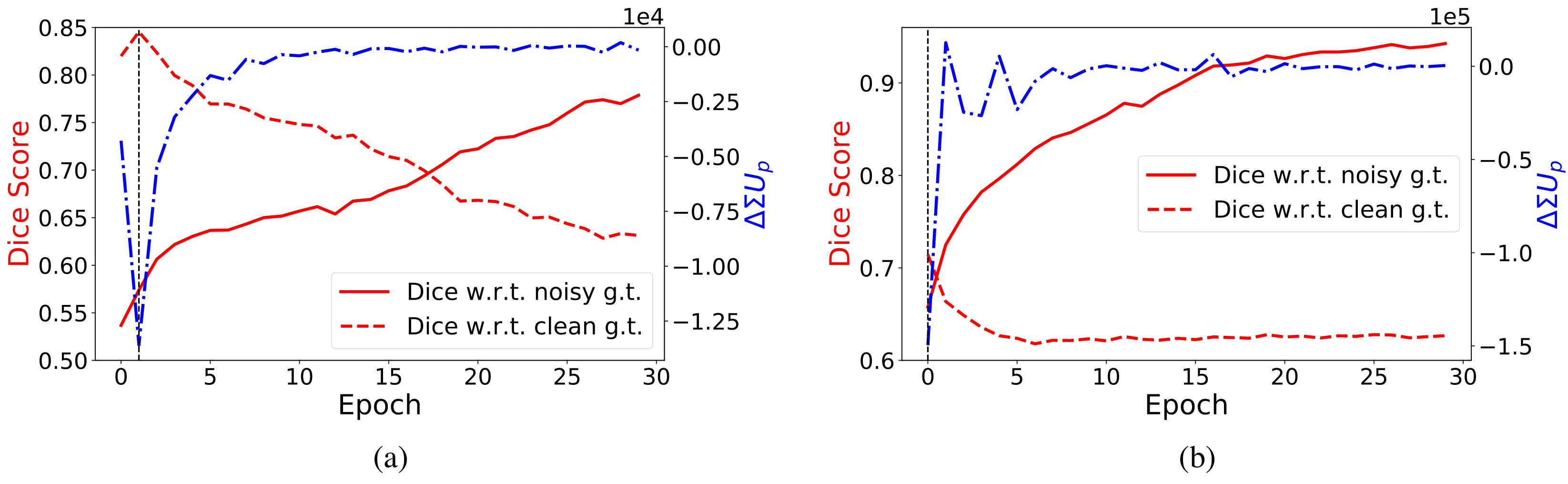}
\caption{Dice score of segmentations produced by the DCNN computed with respect to noisy and clean masks, and relative change of mean cumulative uncertainty for: (a) Dermatoscopy and (b) Liver CT datasets\label{fig:noisy_metrics}} 
\end{figure*}



Many studies on uncertainty estimation address the Bayesian modeling of DCNNs. 
Monte Carlo Dropout (MCDO) \cite{gal2016dropout} which has been successfully applied in medical imaging \cite{kwon2018uncertainty} aims to approximate (in the Bayesian sense) a probabilistic Gaussian process at inference time, by running the trained model $N$ times with activated dropout layers. Deep Ensembles (DE) were proposed as a non-Bayesian approach for predictive uncertainty estimation \cite{lakshminarayanan2017simple}. The method is based on constructing an ensemble of models with similar architectures and random initializations, and consider the variance of predictions as a measure of uncertainty. Recently, test-time augmentations (TTA) were proposed as a mechanism of uncertainty estimation \cite{wang2019aleatoric}. This approach uses geometric transformations of the input (flipping, rotation, scaling, etc.) to obtain an ensemble of predictions, the variance of which can represent uncertainty. 

The authors of \cite{kendall2017uncertainties} proposed to consider two types of predictive uncertainty, $\it{epistemic}$  and $\it{aleatoric}$. Epistemic ($\it{model}$) uncertainty can be explained away given enough training data. Aleatoric ($\it{data}$) uncertainty depends strictly on the noise in the input images, and can therefore be used to detect and relabel wrongly labeled pixels. 
Considering the case of binary segmentation of images of size $H\times W$ and applying one of the uncertainty estimation methods, the resulting tensor of pixel labels predictions $\boldsymbol{\hat{p}}$ has shape $N\times H\times W$. Then, according to \cite{kwon2018uncertainty}, the aleatoric uncertainty can be calculated as follows:

\begin{equation}
U_{p} = \frac{1}{N}\sum_{n=1}^{N}\text{diag}(\boldsymbol{\hat{p}_{n}}) - \boldsymbol{\hat{p}_{n}\hat{p}_{n}^{T}}, \label{eq:unc}
\end{equation}

\subsection{Detection and relabeling of noisy pixel labels}
\label{ssec:subhead2}
The areas of high uncertainty are the ones for which the predictions generated by one of the methods described above demonstrate more variability. Pixels in these areas are likely to have incorrect labels. In case of binary image segmentation, such labels can be easily flipped according to this rule: $p_{new} = 1-p_{old}$,  if $U_{p}>\delta$, where $p_{old} \in {0, 1}$ is the label of the pixel in the ground truth mask, and $\delta$ is a predefined threshold on uncertainty.




The relabeling procedure consists of three steps shown in Figure~\ref{fig:pipeline2}. First, regions of high and low uncertainty are detected on the uncertainty map computed using (\ref{eq:unc}). Then, according to the flipping rule, each pixel in the noisy mask changes its value to the opposite one if it lies in the area of high uncertainty, and keeps its current label otherwise. Finally, since holes may appear in the masks after relabeling, that are due to uncertainty map values fluctuating near the threshold value, morphological hole filling may be applied to the newly generated masks. 

\begin{figure*}[!htb]
\centering
\includegraphics[width=1\textwidth]
{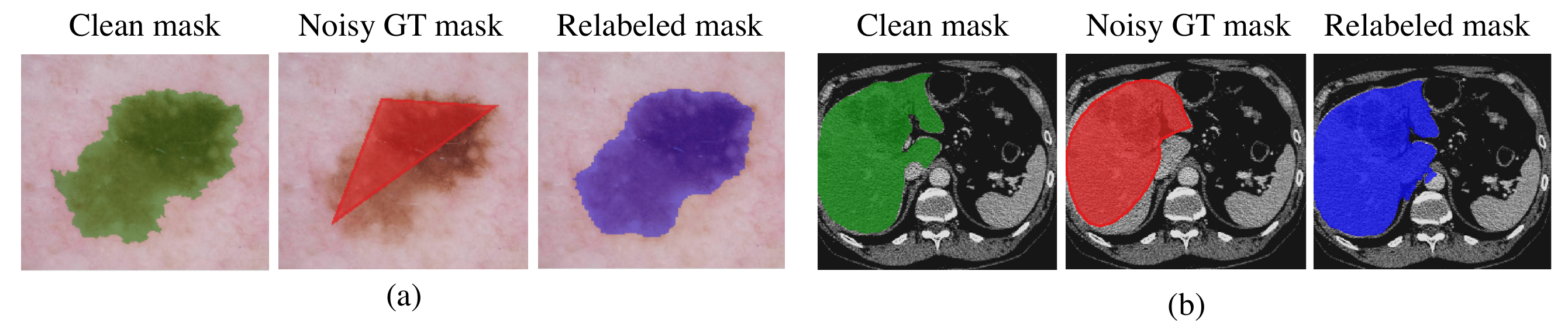}
\caption{Relabeling results for (a) Dermatoscopy dataset, (b) Liver CT dataset\label{fig:rel_res}} 
\end{figure*}

\section{Experiments and Discussion}
\label{sec:ExpDis}

\subsection{Image segmentation with noisy labels}
\label{ssec:seg}
The proposed solution was validated on two datasets. The first one is provided by the International Skin Imaging Collaboration \cite{codella2018skin} and consists of 2000 training and 150 validation dermatoscopic images of skin lesions. The second is a liver segmentation dataset provided by the organizers of Medical Segmentation Decathlon \cite{simpson2019large}. The data consists of 201 contrast-enhanced CT images and was divided into train and test in the same proportions as in the first dataset. For both datasets, segmentation masks for two classes are provided. We assumed that both datasets consisted of only clean annotations, and generated noisy ground truth masks as described in section 2.1.

In the first experiment we compared our uncertainty-based relabeling method to another work on segmentation with noisy ground truth labels~\cite{mirikharaji2019learning} that assigns a weight for every pixel in the image. Results are presented in Table~\ref{liver_comp}. In each column, the segmentation accuracy of a trained DCNN is represented by two numbers, $\text{D}_{clean}$ and $\text{D}_{noisy}$: the Dice score, in percent, computed at test time with respect to clean masks and noisy masks correspondingly. The results indicate that when there is no correction of noisy masks, the neural network overfits to the noisy data, resulting in $\text{D}_{noisy} > \text{D}_{clean}$. Reweighting and relabeling approaches both solve this problem and lead to higher $\text{D}_{clean}$. The effect is more pronounced for the dermatoscopy dataset in case of very crude masks. Before we applied noisy mask correction to this data, $\text{D}_{clean}$ was equal to 57.9\% and $\text{D}_{noisy}$ was 65.2\%. The application of our approach resulted in $\text{D}_{clean}=79.4\%$ and $\text{D}_{noisy}=49.3\%$. This indicates that the training of a binary segmentation DCNN switched from overfitting to noisy data towards learning the parameters that would produce segmentation masks that are more similar to clean ones.    

In terms of segmentation accuracy, our method is on par with the reweighting approach from \cite{mirikharaji2019learning} for both datasets. The only exception is the case of triangular liver masks in CT images where our method performs much worse. This is due to the fact that overfitting to noisy data happens very quickly during the training. We will discuss this phenomenon next.
Compared to prior art in~\cite{mirikharaji2019learning}, the main advantage of our method is that only noisy masks are required for its applicability, while the reweighting method requires some amount of masks that are known to be completely noiseless.


\begin{table}[]
\resizebox{\columnwidth}{!}{%
\begin{tabular}{lcccc}
\hline
\multicolumn{5}{c}{\textit{\textbf{Dermatoscopy dataset}}}                                                                                                      \\ \hline
\text{Clean GT}               & \multicolumn{4}{c}{82.5}                                                                                                          \\ \hline
\multicolumn{2}{l}{}                           & \multicolumn{1}{l}{\text{3 vertices}} & \multicolumn{1}{l}{\text{7 vertices}} & \multicolumn{1}{l}{\text{Smooth}} \\ \hline
\multicolumn{2}{l}{No noise correction}        & 57.9 / 65.2                           & 78.3 / 80.2                           & 78.7 / 80.6                         \\
\multicolumn{2}{l}{Reweighting \cite{mirikharaji2019learning}}                & 79.5 / 55.4                                 & 80.7 / 79.8                                 & 80.3 / 79.3                               \\
\multicolumn{2}{l}{\text{Relabeling (ours)}} & 79.4 / 49.3                           & 80.5 / 79.7                           &  \textbf{81.5 / 80.4}          \\ \hline
\multicolumn{5}{c}{\textit{\textbf{Liver CT dataset}}}                                                                                                               \\ \hline
\text{Clean GT}               & \multicolumn{4}{c}{94.4}                                                                                                          \\ \hline
\multicolumn{2}{l}{}                           & \multicolumn{1}{l}{\text{3 vertices}} & \multicolumn{1}{l}{\text{7 vertices}} & \multicolumn{1}{l}{\text{Smooth}} \\ \hline
\multicolumn{2}{l}{No noise correction}        & 61.5 / 69.5                                & 85.3 / 86.2                                 & 85.1 / 86.3                               \\
\multicolumn{2}{l}{Reweighting \cite{mirikharaji2019learning}}                & 84.6 / 62.1                                & 87.2 / 85.9                                 & 87.2 / 86.1                             \\
\multicolumn{2}{l}{Relabeling (ours)}          & 72.2 / 63.3                                 &  \textbf{88.0 / 87.1}                 & 87.6 / 85.6                               \\ \hline
\end{tabular}
}
\caption{Dice score, in percent, with respect to clean / noisy masks, of the same segmentation DCNN trained using simulated noisy data, with and without noise correction. Dice score for training on original clean data is provided as a reference. See section \ref{ssec:seg} for details.
\label{liver_comp}}
\end{table}

\subsection{Relabeling strategies}
\label{ssec:rel_str}
During training, it is essential to start the relabeling process before the network overfits to noisy data. In Figure~\ref{fig:noisy_metrics} we show learning curves for the dermatoscopy and liver CT datasets with noisy masks approximated by low-vertex polygons. For dermatoscopy data, the network has learned enough to correctly relabel the noisy ground truth data around the second epoch of training, as $\text{D}_{clean}$ (the red dashed curve) starts to drop after reaching a peak. The learning curve for liver CT data shows that the relabeling should start right after the first epoch before it overfits to noisy masks. Of course, this information is not available during training, if the provided ground truth masks are noisy. In this case, there are two possible solutions to identify the moment when the predictions should be used for data relabeling. One is to use a small subset of training data, handle it to experts for accurate relabeling, and train the network on the noisy version and test on the clean one. It was shown in~\cite{kohler2019uncertainty} that it is not required to know the exact true share of noisy images in the dataset in order to correctly identify the epoch to perform relabeling. The second solution does not require any knowledge about what data samples are clean and relies on the analysis of uncertainty estimates over the course of training. For this analysis, we calculated the cumulative uncertainty $U_{p}$ using formula~(\ref{eq:unc}), summed it over all the pixels of an image and averaged over data samples $\Sigma U_{p}$. We then examined its relative change $\Delta\Sigma U_{p}$ during the course of training. 

The graph of $\Delta\Sigma U_{p}$ is shown in Figure~\ref{fig:noisy_metrics} (a) and (b) by the blue curves. We argue that in the absence of data that is guaranteed to be clean, a possible indicator of the epoch suitable for relabeling could be the point when the speed at which the mean cumulative uncertainty decreases is at its maximum (in other words, when $\Delta\Sigma U_{p}$ reaches its minimum). These epochs are indicated by vertical lines in Figure~\ref{fig:noisy_metrics}. At this moment, the network has already learned enough to be highly uncertain mostly in the regions of noisy boundaries. When the network starts overfitting to noise, $\Delta\Sigma U_{p}$ goes to zero. For both datasets, the epoch indicated by this minimum in $\Delta\Sigma U_{p}$ coincides with the epoch when the segmentation accuracy measured with respect to the clean test data reaches a maximum before dropping if there is no noise correction. 

In Figure~\ref{fig:rel_res} we show the masks that were relabeled after the second epoch for dermatoscopy dataset and after the first epoch for liver CT dataset, while training a binary segmentation DCNN using noisy GT masks. In both cases, the relabeled masks are very similar to the clean masks, despite they were never used during training.

\subsection{The effect of uncertainty estimation methods}
\label{ssec:unc_meas}
We studied the effect of the three methods of uncertainty estimation described in section \ref{ssec:sub_unc} for relabeling noisy masks (3-vertex polygons) from the dermatoscopy dataset. We obtained very similar Dice scores with respect to clean masks in the test set, namely $\text{D}_{clean}=79\%$ for the three methods. This means that the uncertainty estimation method should be chosen by criteria other than the final segmentation quality, e.g. by analyzing computational complexity. Unlike MCDO, the DE and TTA methods do not require network modification. On the downside for DE, the model requires retraining from scratch, which is computationally expensive for large datasets. In our opinion, TTA seems to be the least expensive approach as it requires only applying transformations to data samples without retraining or modifying the segmentation network.

\section{Conclusions}
\label{sec:Concl}
We have demonstrated that predictive uncertainty can be used for detecting noisy pixel-level labels in inaccurately annotated ground truth segmentation masks. We did so by applying several uncertainty estimation methods and verified the relabeling of two image segmentation datasets. Automatic relabeling improves segmentation quality without overfitting to data inaccuracies. The proposed algorithm can be used to generate cleaner datasets for training other deep learning algorithms without having to consider the impact of noisy labels.
\newpage

\section{Compliance with Ethical Standards}
This research study was conducted retrospectively using human subject data made available in open access by the International Skin Imaging Collaboration and by the organizers of Medical Segmentation Decathlon. Ethical approval was not required as confirmed by the license attached with the open access data.

\section{Acknowledgments}
The authors declare no conflicts of interest.


\bibliographystyle{IEEEbib}

\bibliography{refs}

\end{document}